\documentclass{article}
\usepackage{arxiv}
\usepackage{adjustbox}
\usepackage{blindtext}
\usepackage[utf8]{inputenc} 
\usepackage[T1]{fontenc}    
\usepackage{hyperref}       
\usepackage{url}            
\usepackage{booktabs}       
\usepackage{amsfonts}       
\usepackage{nicefrac}       
\usepackage{lipsum}		
\usepackage{graphicx}
\usepackage{doi}
\usepackage[table, svgnames]{xcolor}
\usepackage{multirow}
\usepackage{geometry}
\usepackage{float}
\usepackage[table, svgnames]{xcolor}
\usepackage{tabularx, caption, makecell, booktabs,enumitem}
\usepackage[utf8]{inputenc}
\usepackage{tikz,float}
\usepackage{tabularx}
\usepackage{xparse}

\usepackage{amsmath,amssymb,amsthm}
\usepackage{makecell,rotating}
\usepackage[justification=centering]{caption}
\usepackage{array}

\title{A Survey on Human Action Recognition }

\date{} 					
\author{
  Shuchang Zhou \\
 University of Electronic Science and Technology of China \\
   \texttt{202222011839@std.uestc.edu.cn} \\
}
\NewDocumentCommand{\statcirc}{ O{#2} m }{%
    \begin{tikzpicture}
    \fill[#2] (0,0) circle (1.0ex); 
    \fill[#1] (0,0) -- (180:1ex) arc (180:0:1ex) -- cycle; 
    \end{tikzpicture}
}

\renewcommand{\headeright}{}
\renewcommand{\undertitle}{}
\renewcommand{\shorttitle}{\textit{} }
\hypersetup{
pdftitle={A template for the arxiv style},
pdfsubject={q-bio.NC, q-bio.QM},
pdfauthor={David S.~Hippocampus, Elias D.~Striatum},
pdfkeywords={First keyword, Second keyword, More},
}
\begin{document}

\maketitle

\begin{abstract}
Human Action Recognition (HAR), one of the most important tasks in computer vision, has developed rapidly in the past decade and has a wide range of applications in health monitoring, intelligent surveillance, virtual reality, human-computer interaction and so on. Human actions can be represented by a wide variety of modalities, such as RGB-D cameras, audio, inertial sensors, etc. Consequently, in addition to the mainstream single modality-based HAR approaches, more and more research is devoted to the multimodal domain due to the complementary properties between multimodal data. In this paper, we present a survey of HAR methods in recent years according to the different input modalities. Meanwhile, considering that most of the recent surveys on HAR focus on the third perspective, while this survey aims to provide a more comprehensive introduction to HAR novices and researchers, we therefore also investigate the actions recognition methods from the first perspective in recent years. Finally, we give a brief introduction about the benchmark HAR datasets and show the performance comparison of different methods on these datasets.

\end{abstract}


\section{Introduction}
As an important part of computer vision, HAR is a key technology for machines to understand the world as well as human behavior. In recent years, with the continuous development of deep learning and sensor technology, the performance of HAR has been significantly improved, and it has a wide range of practical applications, including health-care, human-computer interaction, virtual reality, etc. 

Reviewing the development of HAR, a large number of research was based on RGB at the beginning. At the early stage, extensive studies focusing on hand-crafted feature-based approaches have been carried out \cite{1,2,3,4}. With the rise of deep learning, a growing number of neural network-based methods have been proposed with excellent performance, which are categorized into blocks, including two-stream  Convolutional Neural Network (CNN) \cite{5,6,7,8,9,10,23}, Recurrent Neural Network (RNN) \cite{11,14,15,13,18,24,25,26}, 3D CNN \cite{27,28,31,33,29}, and Transformer-based methods \cite{35,37,36,41,42,40}. In recent years, due to the emergence of various sensors, human actions can be represented by other diverse modalities.3D skeleton modality has attracted the interest of many researchers, and skeleton-based HAR using deep learning framework for skeleton sequence perform well, which can be mainly divided into four categories: RNN \cite{49,50,51,52,53}, CNN \cite{55,54,56,57},) and graph convolution network (GCN) \cite{59,60,61,62,63,64,65,66,67}. 

Since single mode has its own limitations, the concept of multi-modality has been proposed and attracted great attention. Multimodal-based methods can further improve the performance of action recognition by taking advantage of the complementary characteristics between multimodal data. Compared with the action recognition based on single modality, the multimodal-based HAR has higher accuracy. However, it may bring various challenges such as data acquisition, feature extraction, fusion, and temporal synchronization. From the traditional perspective, this paper focuses on the multimodal fusion between visual sensors and non-visual sensors, including fusion of RGB and audio modality\cite{74,75,72,78,81}, fusion of RGB and inertial sensors modality \cite{82,83,84,85}, and fusion of RGB-D and inertial sensors modality\cite{91,94,95,97,98,99,100,101,102}. 

At the same time, in recent years, thanks to the popularity of wearable intelligent devices and the rapid development of video social platforms, there is a blowout type growth of video data under the first perspective, and egocentric action recognition(EAR) has become an active research area. The difference between first person point of view (FPV) and a third person view is that the former has a certain initiative, and it is closer to the human eye perception, namely, the camera’s angle is driven by people's attention itself, while conventional perspective camera is fixed. In that case, the action recognition methods based on FPV and a third person view have certain differences, EAR is worth further exploration. Currently, EAR has been applied in various fields, including extreme sports, health monitoring, life recording, etc., and more and more researchers are exploring in the field of EAR \cite{108,109,112,119,122}. In the past few years, several features based on egocentric cues\cite{108,110,112,113,114} such as gaze, hand and head movements, and hand posture have been suggested for first-person action recognition. With the rise of deep learning, advanced deep learning-based methods for EAR are increasingly available. The networks used for extracting spatial-temporal information from self-centered videos can be divided into two categories. The first category \cite{118,119,120,121,123} is based on CNN networks to generate spatial-temporal features, and the second category \cite{124, 35, 126, 127} mainly uses Long Short-Term Memory (LSTM) and its variants. However, most of these methods require large amounts of annotation data. EAR has been further investigated by reweighting spatial or temporal features through Attention Mechanism \cite{128,129,125,126,130}. In addition, joint modeling studies on egocentric gaze prediction and action recognition \cite{131,132,133,134} have addressed the uncertainty in gaze. What is more, there are some other approaches from recent years \cite{135,136,137}. 

In spite of the fact that there are few methods for multimodal action recognition from the first perspective, they have already attracted the attention of researchers. This paper briefly introduces some viable multimodal fusion approaches for EAR, which utilizes data from RGB and depth \cite{138,139,142}, RGB and audio \cite{146,74,148,149,150,151}, RGB and inertial sensors \cite{153,154,155,156,157,158,159}.

\begin{figure}[h]
\centering
\includegraphics[width=0.8\textwidth]{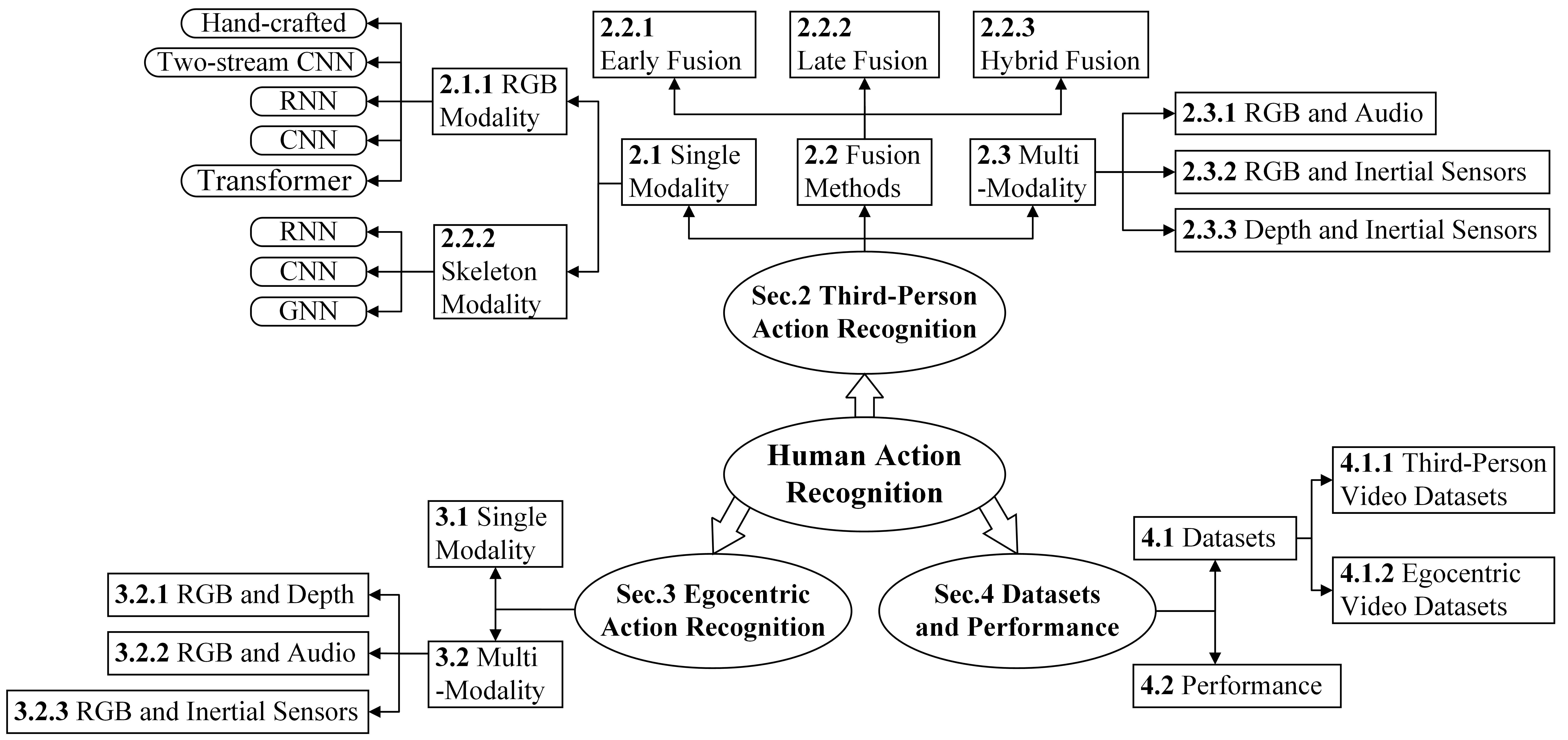} 
\caption{Structure of this paper.}
\label{fig:structure}
\end{figure}

The contributions of this survey can be summarized as follows:
\begin{itemize}
 \item we classify and sort the third-person HAR methods in detail according to different learning frameworks. Meanwhile, multimodal action recognition methods from the third perspective are summarized as well.
 \item we investigate the egocentric action recognition methods based on single modality and multi-modality, which just came up in the last few years.
 \item we introduce a variety of widely used action recognition datasets, meanwhile, performance of representative action recognition methods are summarized and compared. This survey provides some guidance for researchers who are interested in this direction.
\end{itemize}
The remainder of this survey is organized as follows. In Section \ref{Third-Person Action Recognition}, for the third perspective, single-modal HAR based on RGB and skeleton data, multimodal fusion methods, and multimodal HAR are reviewed respectively. Section \ref{Egocentric Action Recognition} introduces the egocentric action recognition based on single modality and multi-modality. Section \ref{Datasets and Performance} briefly introduces the benchmark datasets of action recognition and shows the performance of representative methods on these datasets. The structure of the paper is shown in Figure \ref{fig:structure}.

\section{Third-Person Action Recognition}\label{Third-Person Action Recognition}
\subsection{Single Modality-Based HAR }
In the field of HAR, methods based on single modality are soon going to reach maturity. Different single modal data including RGB, skeleton, depth, infrared, audio, inertia sensors, etc. have their own characteristics and many researches have been conducted on action recognition based on different modalities \cite{8,64,78,73,76,88}. This paper mainly focuses on the two mainstream single modality behavior recognition, which are based on RGB data and 3D skeleton data, respectively.\par
\subsubsection{RGB Modality}
\begin{figure}[h]
\centering
\includegraphics[width=0.5\textwidth]{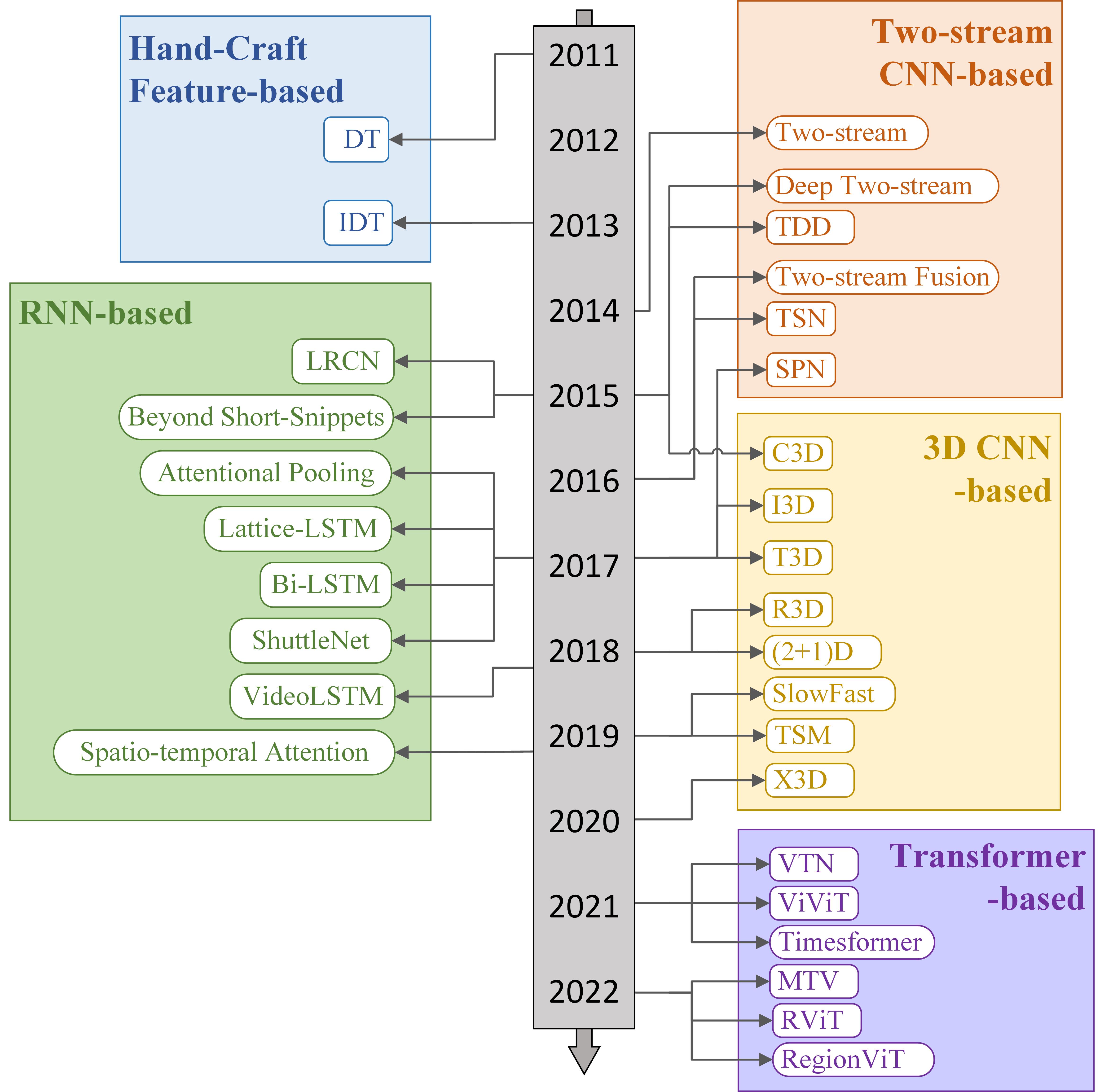} 
\caption{Representative third-person HAR methods based on RGB modality.}
\label{fig:RGB}
\end{figure}

 Looking back at the development of HAR, a large number of studies are conducted based on RGB data, and among which the representative third-person HAR methods are shown in the figure \ref{fig:RGB}. In the first few years, extensive studies focusing on hand-crafted feature-based approaches have been proposed. With more advanced computers and networks, the proliferation of video data, as well as the rapid development of deep learning, a growing number of studies of HAR is conducted based on the deep learning framework. The results show that the HAR methods under the deep learning have superior performance, which gradually replace the traditional methods and have become the mainstream research. These representative deep learning frameworks are elaborated later, including two-stream CNN, RNN, 3D CNN, and Transformer.\par
 
\textbf{Hand-crafted feature methods}.
Before the rise of deep learning, researchers used traditional recognition methods, i.e., manually extracting features through machine learning and classifying them by relevant algorithms. There are three main categories, spatio-temporal volume based methods \cite{1}, spatio-temporal interest point (STIP) based methods  \cite{2} and trajectory based methods \cite{3,4}. Bobicket et al. \cite{1} proposed to use the motion-energy images (MEI) and motion-history images (MHI) by projecting the human body along the time axis in a 3D cube, then classify behavior through template matching method. However, this method is not applicable to complex scenes. The classical STIP method was mainly proposed by Laptev \cite{2}, whose main idea is to extend the feature detection technology from 2D image to 3D spatio-temporal and calculate its feature descriptors, but this method ignores many details of video, and has weak generalization ability as well. Wang et al. \cite{3,4} proposed the dense trajectory (DT) and improved trajectory (IDT). Spatial feature points are detected on each frame of image, and these feature points are tracked individually at each scale to form a trajectory of fixed length, which is finally described by descriptors. The advantage of IDT method lies in the estimation of camera motion by matching SURF descriptors and dense optical flow feature points between the front and back frames, thus eliminating the effect caused by camera motion. It is classified finally by a support vector machine after feature extraction. DT and IDT, although being traditional methods, are comparable to some deep learning methods, meanwhile the method can achieve fantastic results when combined with deep learning framework. The disadvantage is the speed of this algorithm is slow and it needs to accurately track feature points, which is challenging for computers.\par

\textbf{Two-stream CNN-based methods}.
Simonyan et al. \cite{5} proposed a two-stream CNN model composed of spatial flow and temporal flow, where spatial flow obtains appearance information and the temporal flow obtains motion information, and finally, the average method or SVM is used for fusion of classification scores. Subsequently, quite a few methods have been proposed based on the improvement of the two-stream model \cite{6,7,8,9,10,23}. Wang et al. \cite{6} found that many two-stream CNNS are relatively shallow, therefore, they designed very deep two-stream CNNS to obtain better recognition results. Feichtenhofer et al. \cite{8} studied several fusion strategies and proposed that it is effective to fuse spatial flow and temporal flow in the last convolutional layer, thus reducing the number of parameters while maintaining accuracy. In order to solve the problem of not being able to model long time domain structures, Wang et al. \cite{7} proposed temporal segment networks (TSN), which design a sparse sampling scheme to represent temporal features that could model the whole video. The authors proposed two additional input modalities: RGB difference and warped optical flow fields to improve the learning efficiency of the original two-stream network. Wang et al. \cite{9} proposed trajectory-pooled deep-convolutional descriptors (TDD) by combining classical IDT manual features and two-stream depth features. A novel spatiotemporal pyramid network (SPN) \cite{23} was proposed to integrate the spatiotemporal features in the pyramid structure so that they can enhance each other. Peng et al. \cite{10} proposed the two-stream collaborative learning with spatial-temporal attention (TCLSTA) approach, which consists of the spatial-temporal attention model and the static-motion collaborative model, consequently improves the recognition performance by exploiting the strong complementarity of static and dynamic information.
Two-stream CNN networks are able to capture high semantic representations by using CNNS on spatiotemporal features. However, precomputed optical flow is computationally expensive and storage demanding, meanwhile, the networks have limitations in modeling the video-level temporal information effectively as it is insensitive to the time series information in the video.\par

\textbf{RNN-based methods}.
The time series information in video data, which HAR needs to learn from videos, is one of the important factors. Therefore, RNN \cite{11,14,15,13,18} is an ideal choice. Among them, the research areas in video-based HAR using LSTM \cite{17} have attracted a lot of attention. In LRCN \cite{11}, CNN features are extracted from a single frame and then input to LSTM for HAR task. Beyond Short-Snippets \cite{12} extract features from pre-trained 2D CNNS and feed these features to the stacked LSTM framework. \cite{14,15} adopted bidirectional LSTM, which is composed of two independent LSTMS for learning the forward and backward temporal information of HAR. Sun et al. proposed Latch-LSTM \cite{13} to learn the independent hidden state transition of LSTM memory units, so as to further effectively model the dynamic information in video sequences for better classification. In addition to using LSTM, some studies have conducted HAR via GRU \cite{18,19,20,21,22}, which has fewer gates \cite{16} compared to LSTM, resulting in fewer model parameters, but it can generally provide similar performance to LSTM for HAR. ShuttleNet \cite{21} is a deep network inspired by biology that is embedded in CNN-RNN framework containing a multilayer loop-connected GRU processor. In addition, there are some studies that incorporate attention mechanisms \cite{24, 25, 26} to benefit LSTM-based framework to obtain better HAR performance.\par
\textbf{3D CNN-based methods}.
Recently, 3D CNN-based methods \cite{27,28,31,33,29} have achieved good performance in HAR. C3D \cite{27} is one of the earliest video-based 3D CNN models for HAR, where video frames with spatial and temporal dimensions are directly fed into a 3D CNN without any preprocessing. I3D \cite{28} combines two-stream network and 3D CNN to extend 2D Inception-v1 network to 3D structure. Given the ability of ResNet to alleviate the degradation problem of network deepening, Tran et al. \cite{31}designed 3D Residual Networks (R3D), which extend 2D convolution of ResNet to 3D. Tran et al. \cite{33} proposed to use (2+1)D convolution instead of 3D convolution, decomposing the 3D convolution operation into two-dimensional spatial convolution and one-dimensional temporal convolution. T3D  \cite{34} can intensively and effectively capture temporal information. SlowFast network  \cite{32} contains two 3D CNN networks to handle slow and fast paths running at different frame rates. Lin \cite{30} added Temporal Shift Module (TSM) to the ResNet to achieve the performance of 3D CNN, while keeping the relatively small computation cost of 2D CNN. X3D  \cite{29}  attempts to expand 2D convolution from an different dimension to make it suitable for 3D spatio-temporal data processing.\par
\textbf{Transformer-based methods}.
Transformer  \cite{35} with attention mechanism as the core is a novel research hotspot recently, due to its powerful ability and broad prospects, and the application of transformer to RGB-based action recognition has achieved unprecedently superior performance  \cite{36,41,42,39,40,neimark2021video}. Bertasius et al.  \cite{36} extended ViT (ViT  \cite{37} is another Transformer used for image classification) to videos by decomposing each video into a series of frame-level patches, then, a dividing attention mechanism is proposed to apply spatial and temporal attention separately within each block of the model. ViViT  \cite{41} extracts spatio-temporal feature information from the input video, which is then encoded by a series of transformer layers. MTV-H (WTS)  \cite{42} introduces multi-view converters consisting of multiple individual encoders, which effectively fuse information from different representations of the input video. However, they also suffer from severe memory and computation overhead. Therefore, many efforts have been made to reduce the computational complexity and memory cost such as RegionViT  \cite{39}, RViT  \cite{40}.
\subsubsection{3D Skeleton Modality}
\begin{figure}[h]
\centering
\includegraphics[width=0.5\textwidth]{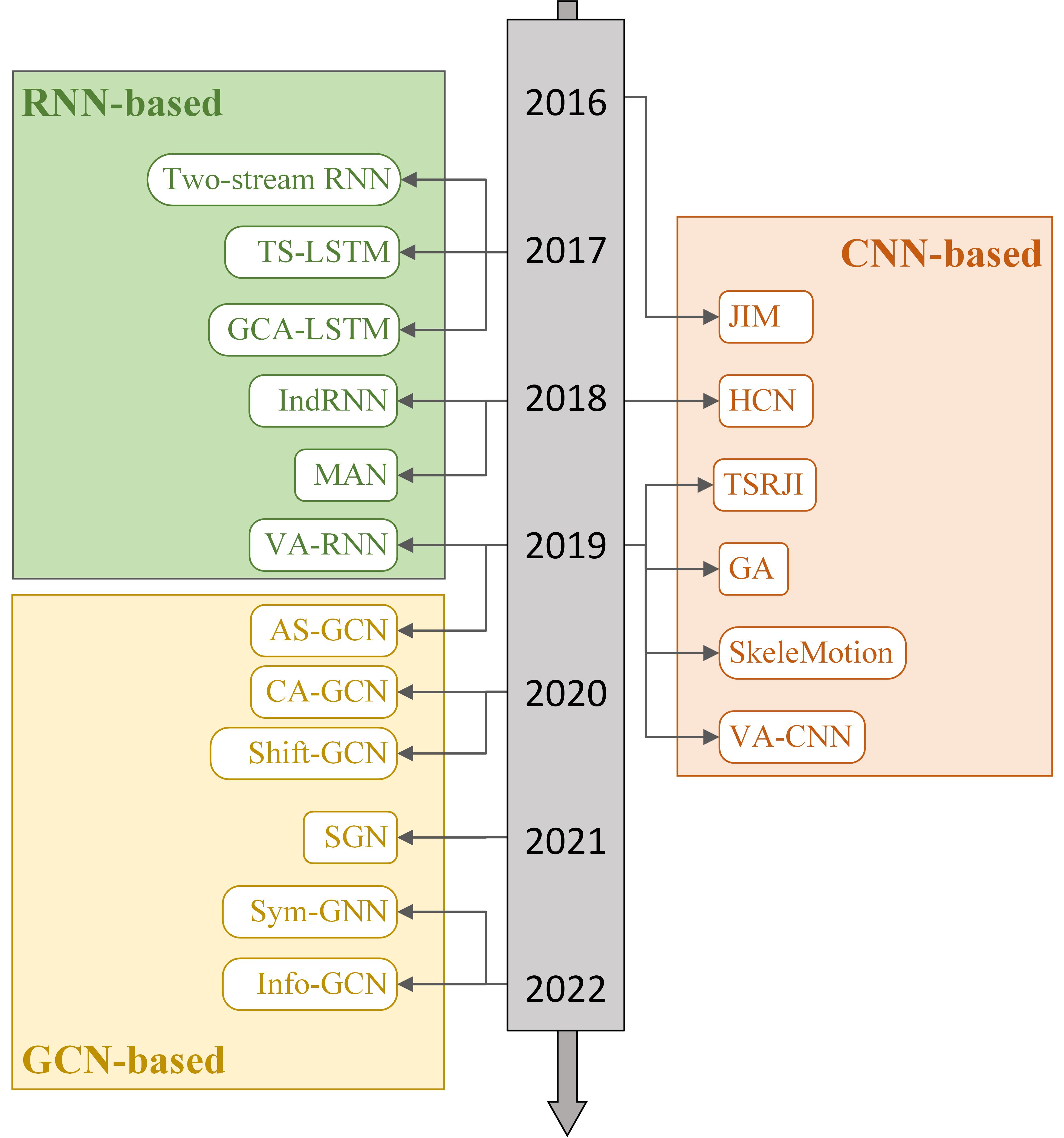} 
\caption{Representative third-person HAR methods based on 3D skeleton modality.}
\label{fig:Skeleton}
\end{figure}

3D skeleton data is another common modality as skeletal sequence encodes the trajectory of human joints, which represents informative human motion. Skeletal data therefore has been undergoing a lot of researches in HAR, and the use of skeletons is increasing. In addition, sensors such as Microsoft Kinect  \cite{43} and advanced human pose estimation algorithms  \cite{44,45} also make it easier to obtain accurate 3D skeleton data.
 The earliest skeletal sequence-based action recognition use hand-crafted feature-based methods  \cite{46,47,48}. However, these traditional methods can only perform well in specific datasets, it is difficult for these methods to be applied in a wider application fields. With the rise of deep learning, skeleton sequence HAR based on deep learning framework has gradually become the mainstream. In the following, we review deep learning methods, which can be mainly divided into four categories: RNN, CNN, Graph Convolution Network (GCN), and Transformer-based methods. Representative third-person HAR methods based on 3D skeleton modality are shown in the figure \ref{fig:Skeleton}.\par
 
\textbf{RNN-based methods}.
In the work of  \cite{49}, an end-to-end two-stream RNN architecture was proposed to simulate temporal dynamics and spatial configurations. To model the variable temporal dynamics of skeleton sequences, Lee et al. \cite{53} proposed end-to-end Memory Attention Network (MAN), namely, it is a temporal-then-spatial recalibration scheme designed to alleviate complex spatio-temporal variations of skeleton joints. Lee et al. \cite{50} proposed novel ensemble Temporal Sliding LSTM (TS-LSTM) networks containing short-term, medium-term and long-term TS-LSTM networks, respectively. IndRNN \cite{51} constructs an independent recurrent neural network that not only solves the problems of gradient disappearance and explosion, but also faster than the original LSTM. Given that not all joints provide information for action analysis, global context-aware attention \cite{52} is added to the LSTM network, which selectively focuses on the information joints in the skeleton sequence. Zhang et al. \cite{79} proposed adaptive neural networks VA-RNN and VA-CNN, specifically, in each stream, the view adaptive module automatically determines and identifies the best observation point in the process.\par
\textbf{CNN-based methods}.
CNN is obviously better than RNN in image extraction, however, its problem regarding temporal modeling is still the direction of focus. Wang \cite{54} proposed Joint Trajectory Graph (JTM), which represents the spatio-temporal information in 3D skeleton sequences into three 2D images through color coding. Since only adjacent joints within the convolution kernel will be considered to learn co-occurrence features, that is, some potential correlations associated with all joints may be ignored. Li et al. \cite{55} proposed an end-to-end framework for learning co-occurrence features using a hierarchical approach. A geometric algebraic representation \cite{56} of shape motion was proposed, which makes full use of the information provided by skeletal sequences. Similarly, Caetano et al. proposed SkeleMotion \cite{57} and the Tree Structure Reference Joints Image (TSRJI) \cite{58} as representations of HAR skeleton sequences.\par
\textbf{GCN-based methods}.
Given that skeletal sequences based on CNN or RNN cannot completely simulate the spatio-temporal information, while the skeleton data is in the form of graph structure, which also has strong expressive ability. Inspired by topological graph, HAR methods based on GNN or GCN have been proposed successively, and their experimental results show that the graph structure viewing skeleton data as edges and nodes can better carry out HAR. Since GCN methods have been extensively studied, this paper will focus on GCN-based HAR.

Yan et al.\cite{59} first proposed a novel skeleton-based action recognition model, namely, Spatial Temporal Graph Convolution Network (ST-GCN), which constructs a spatiotemporal graph with joints as graph vertices, meanwhile body structure and time as graph edges. In order to reflect implicit joint correlation, Li et al. \cite{60} further proposed Actional-structural Graph Convolutional Networks (AS-GCN), which combine actional links and structural links into a generalized skeleton graph. Actional links are used to capture potential dependencies specific to actions, while structural links are used to represent higher-order dependencies. Furthermore, in the work of \cite{61}, contextual information integration is used to effectively model long-term dependencies. A high-level semantics of joints is introduced for HAR in \cite{62}. In addition, to reduce the computational cost of GCN, Cheng et al. designed Shift Graph Convolutional Network (Shift-GCN) \cite{63}, which employs Shift-graph operations and lightweight point-wise convolutions instead of using heavy regular graph convolutions. Following this line of research, Song et al. \cite{64} proposed a multi-stream GCN model, specifically, separable convolutional layers are embedded into an early fused Multiple Input Branches (MIB) network where the input branches including joint position, motion velocity, and skeletal features, thus, greatly reduce redundant trainable parameters while increasing the capacity of the model. Li et al. \cite{65} proposed symbiotic GCN to simultaneously handle action recognition and motion prediction tasks, which allows the two tasks to enhance each other. In \cite{66}, InfoGCN was proposed including an information bottleneck objective to learn the most informative action representations, and an attention-based graph convolution to capture context-dependent skeleton topologies.
\subsection{Multimodal Fusion}
Information from different sensing modes can be fused in different ways. Generally speaking, three fusion methods are outlined: early fusion, late fusion, and hybrid fusion.\par
\subsubsection{Early Fusion}
Both feature-level and data-level fusion are referred to as early fusion. Data-level fusion is suitable for homogeneous multi-sensor data (e.g., two or more RGB cameras or depth cameras, etc.). When there are two or more heterogeneous sensors, feature-level fusion or decision-level fusion techniques are usually applied. The modalities are often highly correlated with each other, but it is difficult to extract this correlation in both feature layer and data layer. According to literature \cite{68}, correlation between information contained in different data streams can only be found at a higher level. Researchers usually use dimensionality reduction techniques to eliminate redundancy problems in the input space. For example, the Principal Component Analysis (PCA) method in literature \cite{5} is widely used in dimensionality reduction processing in multimodal deep learning. In addition, the multimodal early fusion method also needs to solve the problem of time synchronization among different data sources, several methods were proposed to solve this problem in \cite{70}, such as Convolutional, Training and Pool Fusion, which can well integrate discrete event sequence and continuous signal to realize time synchronization between modalities.\par
\subsubsection{Late Fusion}
Late fusion is also called decision-level fusion, where deep learning models are trained on different modes first, and then the output results of multiple models are fused. This fusion approach is often favored as the fusion process is feature independent and errors from multiple models are generally uncorrelated. Currently, late fusion methods mainly use rules to determine the how to combination of output results on different models, that is, rule fusion, such as Max-Fusion, Averaged Fusion, Bayes Rule Fusion, and Ensemble Learning, etc. \cite{71}.\par
\subsubsection{Hybrid Fusion}
In \cite{70}, early and late fusion methods are compared and the performance of both methods had a lot to do with specific problems. Namely, early fusion is superior to the late fusion when the correlation between the modes is relatively large, while when each mode is not related to a great extent, such as dimension and the sampling rate are highly uncorrelated, adopting late fusion method is more suitable. Therefore, the two methods have their own advantages and disadvantages, which need to be selected according to the requirements in practical application. Hybrid fusion combines early and late fusion methods, which integrates the advantages of the both, while increases the structure complexity and training difficulty of the model. Due to the diversity and flexibility of deep learning model structures, it is more suitable to use hybrid fusion method, which is widely used in multimedia, gesture recognition and other fields.
\subsection{Multimodal HAR}
Vision sensors include RGB-D cameras, infrared, time of flight, light field, thermal imagers, etc. Visual modalities are widely used because of their excellent representation capabilities. Non-visual sensors include accelerometer, gyroscope, magnetometer, audio signals, electrothermal activity response, etc. The sensors have been used individually or in combination for HAR. The following will mainly focus on multimodal fusion of visual sensors and non-visual sensors, including the multimodal fusion between RGB and audio modalities, RGB and inertial sensors modalities, RGB-D and inertial sensors modalities.\par
\subsubsection{RGB and Audio}
Audio data provides complementary information to appearance and motion in visual data. In recent years, several deep learning-based methods have been proposed to fuse RGB and audio modalities for HAR. The joint modeling of audio and visual signals is mainly carried out in the way of late fusion \cite{74,75}. Wang et al. \cite{72} introduced three-stream CNN to extract multimodal features from audio signals, RGB frames and optical flows. Both feature fusion and score fusion are evaluated, with the former achieving better performance. Inspired by the work of \cite{32}, Xiao et al. \cite{78} introduced a hierarchically integrated audio-visual representation framework with slow and fast visual paths that are deeply combined with multilayers audio paths. Chen et al. \cite{80} proposed a multimodal video convert called Multi-Modal Video Transformer (MM-ViT), which operates in the compressed video domain and exploits all readily available modalities to avoid calculation of optical flow, i.e., I-frames, motion vectors, residuals and audio waveform. In \cite{81}, an audio-infused recognizer was proposed, which effectively models the cross-modal interaction across domains.\par
\subsubsection{RGB and Inertial Sensors}
In addition to vision-based sensors, inertial sensors have been used for human action recognition, allowing recognition beyond the limited field of view of vision-based sensors. Inertial sensors contain accelerometers and gyroscopes that provide acceleration and angular velocity signals for HAR, and a survey \cite{89} details the performance of current deep learning models and future challenges for sensor-based action recognition. The wearable inertial sensors provide 3D motion data, consisting of the three -axis acceleration of the accelerometer and the three-axis angular velocity of the gyroscope. Inertial and video data in \cite{82,83,84} are captured simultaneously by inertial sensors and video cameras and converted into 2D and 3D images. These images are then fed into 2D and 3D CNNs to fuse their decisions in order to detect and identify a specific set of actions of interest from a continuous stream of actions. In \cite{83,84}, a decision-level fusion method is considered. In \cite{82}, both feature-level fusion and decision-level fusion are tested, and decision-level fusion achieves higher accuracy. In \cite{85}, visual and inertial sensor integration algorithms were proposed for efficient and accurate generic abnormal behavior detection among the elderly, which closely cooperate to achieve high accuracy and real-time performance.\par
\subsubsection{RGB-D and Inertial Sensors}
In recent years, many studies have improved the accuracy of HAR by fusing features extracted from depth and inertia sensor data and using co-representation classifiers. Better accuracy results have been obtained due to the complementarity of the data from two modalities. In the vast majority of the work on action or gesture recognition, it is assumed that the action of interest has been separated from the action stream \cite{90,91,92,93,94}. In \cite{91}, decision-level fusion between depth camera data and wearable sensor data is performed to increase the ability of the robot to recognize human actions. In \cite{90}, the depth and inertia data are effectively fused to train the Hidden Markov Model to improve the accuracy and robustness of gesture recognition. For continuous action flow, Dawar et al. \cite{96} detected and recognized human actions from continuous action flow by fusing both depth and inertia sensing modalities.

In addition, many deep learning methods have been proposed recently \cite{97,98,99,100,101,102,89}. In \cite{98}, a shared feature layer is used after multimodal feature-level fusion, and then support vector machines or softmax classifiers are used to recognize actions based on combined features. Considering that deep inertia training data is limited, Dawar et al. \cite{99} proposed a data enhancement framework based on depth and inertia modalities, which are fed to CNN and LSTM, respectively, and then the scores of the two models are fused during testing for better classification. Given that deep learning model allows to extract features at all levels of the structure so that rich multi-layer features are obtained, while existing methods do not take advantage of these rich multi-level information. Specifically, the main drawback of existing deep learning-based HAR fusion methods based on depth and inertia sensors is that the fusion is performed at a single level or stage, either at the feature level or at the decision level, and therefore the true semantic information of the data cannot be mapped to the classifier. By designing different two-stream CNN architectures, several deep inertial fusion techniques are also studied in \cite{101,102}, where inertial signals are converted into images using the techniques in \cite{103}. Three new deep multilevel multimodal ($\emph{M}^{2}$) fusion frameworks are proposed in \cite{101} to take advantage of different fusion strategies at different stages. Ahmad et al. \cite{102} proposed a new Multistage Gated Average Fusion (MGAF) network, which extracts and fuses features from all layers of CNN. Recently, Ijaz et al. \cite{104} proposed a multimodal Transformer for nursing action recognition, in which the correlation between skeleton and acceleration data is modeled by Transformer.

\section{Egocentric Action Recognition}\label{Egocentric Action Recognition}
\subsection{Single Modality-Based EAR}
In the last decade, with the emergence of low-cost and lightweight multi-sensor wearable devices, such as GoPro, Google Glass, Microsoft Hololens, etc., video data from the first perspective has surged year by year, and HAR technology based on first view has been applied to various fields, including extreme sports, health monitoring, life recording and so on. The field of EAR has also published its representative dataset \cite{105,106,107} and has attracted a lot of interest, with more and more researchers have explored the EAR field \cite{108,109,112,119,122} in the past decade. Given that first-person action recognition is not like counterpart based third-view, where the camera is either static or moves smoothly, while in egocentric videos there is a large vibration due to the wearer's head movement. Therefore, it is difficult to apply the third-person action recognition methods directly to the first view.

In the past few years, several features based on egocentric cues \cite{108,110,112,113,114}, such as gaze, movements of hand and head, and hand posture have been suggested for first-person HAR. First of all, \cite{108,109,110,111} recognize the importance of hand in first-person action recognition. Pirsiavash and Ramanan \cite{110} proposed an egocentric behavior representation based on hand-object interaction, developing a combination of HOG features for modeling object and hand appearance during activities. \cite{112,113} find that gaze position is an important cue for EAR, but such fine-grained information is difficult to detect. However, the direct application of local features in egocentric videos is problematic in that it ignores the fact that camera motion is also a useful cue for understanding egocentric behavior. Motion features also play an important role in egocentric action analysis \cite{114,115}, where Ryoo et al. integrated global and local motion information to model interaction-level human activities. In addition, Ying et al. \cite{116} proposed egocentric cues that combine head movement, hand posture, and gaze to better characterize egocentric actions.

With the rise of deep learning, advanced deep learning-based methods for EAR are increasingly available. The networks used to extract spatio-temporal information from egocentric videos can be divided into two main categories, and the first category is based on CNN to generate spatio-temporal features. An Ego-ConvNet combining egocentric cues is proposed in \cite{122}. In \cite{119}, the network is trained to segment the hand and locate the objects, and then cropped the objects to the input of the appearance flow. After that, appearance flow and motion flow are fused at late-level through the fully connected layer to jointly identify objects and action verbs. Zhou et al. \cite{120} proposed a hybrid cascaded end-to-end deep CNN to jointly infer hand maps and manipulate foreground object maps. A two-stream CNN architecture with long-term fusion pools \cite{121} was proposed to efficiently capture the temporal structure of actions with appearance and motion. Wu et al. \cite{123} combined the long-term feature banks containing detection features with 3D CNN to improve the accuracy of target recognition. These approaches mentioned above utilize local information for EAR based on specialized hand segmentation and so on, for which requires a large amount of annotation data and additional information such as hand masks \cite{122, 116} or target information \cite{119} in addition to input images. The second class uses LSTM and variants \cite{124, 35, 126, 127} to generate embedded representations based on temporal relationships between feature frames. In \cite{35}, spatial attention is considered, where significant regions in each frame are taken as the input of LSTM for action recognition, while spatial attention is further correlated through successive frames in \cite{126}, which a two-stage Long Short-Term Attention (LSTA) model is proposed for locating active objects. At the same time, as attention mechanism \cite{128,129,125,126,130} can effectively localize the region of interest on the feature map, video-based EAR is further studied through the attention mechanism. Lu et al. \cite{128} proposed a spatial attention network to predict human gaze in the form of attention maps, which help two streams focus on the most relevant spatial areas in a video frame and thus recognize actions. The method in \cite{129} is a further extension of \cite{128}, which combines gaze information and attention mechanism. Namely, it uses gaze information as a supervision, and then learns a spatial-temporal attention, lastly integrates the corresponding modules into the two-stream model.

Many of the previous work clearly demonstrates the advantages of using egocentric gaze in  HAR based on FPV. However, they all model gaze and action separately rather than jointly. Studies \cite{131,132,133,134} show the joint modeling of egocentric gaze prediction and action recognition. Huang et al. \cite{132} jointly modeled two coupled tasks of gaze prediction and action recognition while taking into account the context information, with two core modules, more specifically, action-based gaze prediction module and gaze-guided action recognition module are proposed. Li \cite{133} proposed a deep model for jointly learning eye gaze prediction and action recognition, which is an extension of \cite{112}, modeling eye gaze information as a probability variable to explain its uncertainty and then obtains the gaze map, which is combined with features extracted by the neural network later on , thereby getting final recognition results.  

Beyond that, recently, Shan et al. \cite{135} developed a hand-object detector to locate moving objects. When the detector is well trained, it can be deployed directly on the target dataset without fine-tuning. An end-to-end interactive prototype learning (IPL) framework \cite{136} was proposed to learn better active object representations by exploiting the motion cues of participants, without the additional cost of object detection and eye gaze annotation. Meanwhile, Plizzari \cite{137} suggested that the intrinsic motion information encoded by the event data is a very valuable modality for EAR.
\subsection{Multimodal EAR}
At present, there are few studies on multimodal fusion based on the first perspective. The following briefly surveys multimodal EAR, which are mainly divided into RGB and depth modalities, RGB and audio modalities, RGB and wearable inertial sensors modalities.\par
\subsubsection{RGB and Depth} 
As described in \cite{138}, the main limitations of RGB video are the lack of 3D information and sensitivity to illumination changes, which the depth modality is able to compensate for. A multi-stream network is proposed in \cite{139} using Cauchy estimator and orthogonal constraints to combine features from RGB, depth and optical flow. The RGB-D egocentric dataset with hand posture annotations is published in \cite{140}, but they do not propose any method based on RGB and depth. However, with the aforementioned methods, they either can only simulate short-term movements or can only consider the temporal structure as the activity proceeds sequentially. In view of this problem, the recently proposed Transformer \cite{35} can be applied to the EAR of RGB-D, Li et al. \cite{142} introduced the Transformer framework for the EAR, where RGB and its corresponding depth data are processed by the interframe attention encoder. whereafter fused by mutual attention blocks.\par
\subsubsection{RGB and Audio} 
Although some work incorporating audiovisual resources have been reported in the first- person action recognition challenge \cite{143,144}, they provide little model detail. Attention mechanisms for action recognition using audio as a modal branch are proposed in \cite{75, 146}, while the use of audio-visual cues for object interaction recognition is still very limited. Inspired by TSN \cite{7}, a Temporal Binding Network (TBN) \cite{74} was proposed, which takes audio, RGB and optical flow as input through three similar CNN network streams, and then uses a Temporal Binding Window (TBW) when fusing audiovisual features. In this method, modalities are fused before time aggregation, with shared modality and fusion weights over time. Simultaneously, the proposed architecture is trained end-to-end, it thus outperforms individual modalities as well as late-fusion of modalities. The model in \cite{148} combines the sparse temporal sampling strategy with the late fusion of audio, spatial and temporal streams, specifically, its audio input is the spectral map extracted from the original audio of the video, while its visual input is RGB and optical stream frames. Wang et al. \cite{149} solved both problems using a technique called gradient blending, which computes the best fusion of modalities according to their overfitting behavior, and demonstrates the importance of audio in EAR area. However, across all publications in the field, there remains an unresolved problem with EAR, that is, the network relies heavily on the environment in which the activity is recorded and does not perform well in unfamiliar environments, which means that the model trained on source labeled datasets does not generalize well to unseen datasets. The ability to generalize to unseen domains is demonstrated in \cite{150} when audio modality combined with RGB, namely, the Relative Norm Alignment (RNA) loss is proposed, which progressively aligns the relative feature norms from the audio and RGB modalities to obtain the domain-invariant features consequently. Multimodal Temporal Context Network (MTCN) \cite{151} was proposed, which learns to attend to surrounding actions and models multimodal temporal context, more specifically, it is constituted by a transformer encoder that utilizes visual and audio as input context and language as output context.\par
\subsubsection{RGB and Wearable Inertial Sensors}  
Ozcan et al. \cite{152} utilized histograms of edge orientations together with the gradient local binary patterns for fall detection, which then combined with three-axis signal of the accelerometer. Experimental results show that the proposed fusion method not only has higher sensitivity, but also significantly reduces the number of false alarms compared with the accelerometer and camera only methods. In \cite{153}, a multi-stream convolutional network is extended to analysis activity in egocentric videos, meanwhile, a novel multi-stream LSTM is proposed for classifying wearable sensor data. Finally, two score fusion techniques, namely average pooling and maximum pooling, are evaluated to obtain recognition results. Song et al. \cite{154} proposed a new technique to integrate temporal information into sensor data with similar trajectories. Moreover, the Fisher Kernel framework is applied to fuse sensor and video data for EAR with Multimodal Fisher Vector (MFV). In the work \cite{155}, features are extracted by applying a sliding window to video and inertial data, Whereafter, using majority voting to fuse the results. For the classification task, methods of Random Forest and Support Vector Machine are taken into consideration. The methods in \cite{156,157} are extensions of \cite{155}, which use time and frequency domain features for acceleration data, and object information encoding hand interaction for visual data. Experiments are carried out on both feature-level fusion and decision-level fusion, and the latter achieves higher accuracy. In \cite{158}, a hierarchical fusion framework based on deep learning is developed and implemented, where LSTM and CNN are used for EAR based on motion sensor data and photo streams at different levels, respectively. Experimental results show that the proposed model enables motion sensor data and photo streams to work in the most suitable classification mode, so as to effectively eliminate the negative impact of sensor differences on the fusion results. A novel framework is proposed in \cite{159}, where multi-kernel learning (MKL) is used to fuse multimodal features in order to adaptively weighs the visual, audio, and sensor features, additionally, feature and kernel weighting and recognition tasks are performed simultaneously. Huang et al. \cite{69} proposed a first-view multimodal framework based on knowledge driven, GCN and LSTM, which improve the performance of EAR under conditions of few samples and ultra-small datasets.
\section{Datasets and Performance}\label{Datasets and Performance}
\subsection{Datasets}
At present, there are a large number of datasets for third-view action recognition, and they are quite perfect. While there is still room for improvement of egocentric datasets, though the video data from the first perspective is growing due to the popularity of wearable devices in the past few years. The following is a brief introduction to the action recognition datasets, which is shown in Table \ref{tab:datasets} at the same time.
\subsubsection{Third-Person Video Datasets}
\textbf{HMDB51}.
Released by Brown university in 2011, HMDB51 \cite{38} is based mostly on movies, but also on public databases and online video repositories such as YouTube. There are a total of 6849 samples, divided into 51 categories, each category contains at least 101 samples.\par
\textbf{UCF101}.
The UCF101\cite{86} dataset was released in 2012. Collected from YouTube, it is an extension of UCF50 \cite{87} and includes 50 original action classes and 51 new classes. The 101 classes can be divided into 5 categories: human-object interaction, human action, human-person interaction, musical instrument playing, and sports, with a total of 13,320 videos.\par
\textbf{Sports-1M}.
Sports-1M \cite{117} consists of 1 million videos annotated with 487 classes. Each class contains about 1000-3000 video clips. Videos in Sports-1M are automatically collected by the YouTube Themes API.\par
\textbf{UTD-MHAD}.
UTD-MHAD \cite{164} consists of dataset collected synchronously in four modes, which include RGB, depth, skeletal position, inertial signals from the Kinect camera and a suite of wearable inertial sensors that can be used for 27 classes of human actions.\par
\textbf{Activitynet}.
Activitynet \cite{160} was launched in 2015. It is a human activity dataset with 200 activity categories and approximately 24K videos, where videos range in length from a few seconds to ten minutes.\par
\textbf{Something-Something}.
Something-Something \cite{161} contains 108,499 videos, spanning 174 tags, each lasting 2-6 seconds. Unlike other datasets that focus on human activities, which are human-object interactions, it is more accurate to say that "hands" interact with objects, such as actions like, drop, and poke.\par
\textbf{NTU RGB+D}.
NTU RGB+D \cite{162} contains 60 kinds of actions, a total of 56,880 samples, which are collected by Microsoft Kinect v2 sensor, and use three different camera angles, in the form of depth, 3D skeleton, RGB and infrared sequence. Two different partitioning criteria are used to divide the dataset into training set and test set, namely, Cross-Subject (C-Subject) and Cross-View (C-View). The Cross-Subject divides the training set and test set according to the person ID, while Cross-View divides the training set and test set by cameras, with samples collected by camera 1 as the test set, and cameras 2 and 3 as the training set.\par
\textbf{NTU-RGB+D 120}.
Recently, an extended version of the original NTU-RGB+D has also been proposed, called NTU-RGB+D 120 \cite{163}, with 120 action classes and 114,480 samples and two protocols of Cross-Subject (C-Subject) and Cross-Setup (C-Setup) similarly.\par
\textbf{Kinetics-400/600/700}.
kinetics datasets, which have been updated annually over the past several years. In 2017, the Kinetics-400 dataset \cite{141} was developed, which contains 400 activity categories and more than 300,000 10-second video clips extracted from YouTube. In 2018, Kinetics-600 \cite{145} was introduced and 100 classes were added. Kinetics-700 \cite{147} was released the following year, adding 100 classes.\par
\subsubsection{Egocentric Video Datasets}
\textbf{ADL}.
The ADL \cite{110} dataset consists of 20 egocentric videos collected by 20 people. Action annotations and object annotations are provided, with a total of 18 action categories and 44 objects annotated.\par
\textbf{GTEA Gaze+}.
The GTEA \cite{112} dataset was performed by 4 different subjects, consisting of 7 long-term activities, 28 videos with a total of 11 action categories, captured using head mounted cameras.\par
\textbf{Dogcentric}.
The Dogcentric \cite{77} dataset, one of the most popular FPV datasets, consists of 209 videos (102 training videos and 107 test videos), with a variety of first-person actions divided into 10 action categories.\par
\textbf{EGTEA Gaze+}.
EGTEA Gaze+ \cite{131} is the largest and most comprehensive FPV action and gaze dataset. Specifically, it contains 86 unique phases of 28 hours cooking activities from 32 subjects, with 10,325 action annotations, 19 verbs, 51 nouns, and 106 unique actions. Moreover, the videos come with audio and gaze tracking, human annotations of actions and hand masks are provided simultaneously.\par
\textbf{EpicKitchens-55/100}.
EpicKitchens-55 \cite{105} was recorded by 32 participants in four cities using head mounted cameras in their native kitchen environments, with 55 hours of video totaling 39,594 action clips. Meanwhile, the action is defined as a combination of verbs and nouns for a total of 125 verb classes and 331 noun classes. EpicKitchens-100 \cite{107}is an extension of Epickitchens-55, which contains 100 hours of 90k action clips, including 97 verb classes and 300 noun classes, recorded by 45 participants in their kitchens in 4 cities.\par

\begin{table}[h]
  \centering
 
  \caption{Summary of representative video action recognition datasets. \\R:RGB,D:Depth, S:Skeleton, Au:Audio, IR:Infrared, I:Inertial.}
   \label{tab:datasets}
    \begin{tabular}{rccccc}
    \toprule
          & \textbf{Dataset} & \textbf{Samples} & \textbf{Catagories} & \textbf{Modality} & \textbf{Year} \\
    \midrule 
    \multicolumn{1}{c}{\multirow{12}*{\shortstack{Third-Person\\ Video\\ Datasets}}} & HMDB51 \cite{38} & 7K    & 51    & R   & 2011 \\
    \multicolumn{1}{c}{} & UCF50 \cite{87} & 7K    & 50    & R   & 2011 \\
    \multicolumn{1}{c}{} & UCF101 \cite{86} & 13K   & 101   & R   & 2012 \\
    \multicolumn{1}{c}{} & Sports-1M \cite{117} & 1.1M  & 487   & R   & 2014 \\
    \multicolumn{1}{c}{} & UTD-MHAD \cite{164}& 861   & 27    & R,D,S,I & 2015 \\ 
    \multicolumn{1}{c}{} & ActivityNet \cite{160}& 27K   & 203   & R   & 2015 \\
    \multicolumn{1}{c}{} & Something-Something \cite{161}& 108K  & 174   & R   & 2017 \\
    \multicolumn{1}{c}{} & NTU RGB+D \cite{162} & 57K   & 60    & R,D,S & 2016 \\
    \multicolumn{1}{c}{} & NTU-RGB+D 120 \cite{163} & 114K  & 120   & R,D,S & 2019 \\
    \multicolumn{1}{c}{} & Kinetics-400 \cite{141} & 306K  & 400   & R   & 2017 \\
    \multicolumn{1}{c}{} & Kinetics-600 \cite{145} & 496K  & 600   & R  & 2018 \\
    \multicolumn{1}{c}{} & Kinetics-700 \cite{147}& 650K  & 700   & R   & 2019 \\
    \midrule 
    \multicolumn{1}{c}{\multirow{6}*{\shortstack{Egocentric\\Video\\  Datasets}}} & ADL \cite{110}   & 436   & 32    & R,I & 2012 \\
    \multicolumn{1}{c}{} & GTEA Gaze+ \cite{112} & 3K    & 42    & R,IR,Au & 2012 \\
    \multicolumn{1}{c}{} & Dogcentric \cite{77} & 209   & 10    & RGB   & 2014 \\
    \multicolumn{1}{c}{} & EGTEA Gaze+ \cite{131}& 10K   & 106   & R,IR,Au & 2018 \\
    \multicolumn{1}{c}{} & EpicKitchens-55 \cite{105}& 40K   & 149   & R,Au & 2018 \\
    \multicolumn{1}{c}{} & EpicKitchens-100 \cite{107}& 90K   & 4053  & R,Au,I & 2020 \\
    \bottomrule
    \end{tabular}%
  \label{tab:addlabel}%
\end{table}%


\subsection{Performance}
With regard of recognition methods from the third-person view, for RGB-based HAR, UCF101, HMDB51, and Kinectis-400 are widely used as benchmark datasets. Table \ref{R} shows the performance of the representative methods on three datasets, which base on RGB modality and multimodal fusion between RGB and audio modality. Since many methods achieve accuracy of more than 97$\%$ on UCF101, Kinetics dataset was introduced to evaluate the accuracy. 

\begin{table}[h]
  \centering
  \caption{Performance of the  methods based on RGB and multimodal fusion between RGB and audio which is occurred by UCF101,HMDB51,Kinetics400 datasets.}
  \label{R}
    \begin{tabular}{ccccccc}
    \toprule 
    \multicolumn{1}{c}{\textbf{Modality}
} &\multicolumn{2}{c}{\textbf{Method}} & \textbf{UCF101}& \textbf{HMDB51} & \textbf{Kinetics400} &  \multicolumn{1}{c}{\textbf{Year}} \\
    \midrule
   \specialrule{0em}{2.5pt}{2.5pt}
    \multicolumn{1}{c}{\multirow{35}{*}{RGB}} & \multicolumn{1}{c}{\multirow{4}{*}{\rotatebox{90}{\textbf{\shortstack{Hand-craft\\feature}}}}} &  Temporal Template \cite{1} &    -   &   -    &    -   & 2001 \\
    \multicolumn{1}{c}{} & \multicolumn{1}{c}{} & STIP \cite{2}  &   -    &   -    &   -    & 2005 \\
    \multicolumn{1}{c}{} & \multicolumn{1}{c}{} & DT \cite{3}    &  -     & 46.60\% &   -    & 2011 \\
    \multicolumn{1}{c}{} & \multicolumn{1}{c}{} & IDT \cite{4}   & 85.90\% & 57.20\% &    -   & 2013 \\
   
  \specialrule{0em}{2.5pt}{2.5pt}
    \cline{2-7}
    \specialrule{0em}{2.5pt}{2.5pt}
    \multicolumn{1}{c}{} & \multicolumn{1}{c}{\multirow{7}[2]{*}{\rotatebox{90}{\textbf{Two-stream  CNN}}}} & Two-stream \cite{5} & 88.00\% & 59.40\% &   -    & 2014 \\
    \multicolumn{1}{c}{} & \multicolumn{1}{c}{} & Deep Two-stream \cite{6} & 91.40\% & 57.20\% &     -  & 2015 \\
    \multicolumn{1}{c}{} & \multicolumn{1}{c}{} & TDD \cite{9}   & 91.50\% & 65.90\% &    -   & 2015 \\
    \multicolumn{1}{c}{} & \multicolumn{1}{c}{} & TSN \cite{7}   & 94.20\% & 69.40\% &   -    & 2016 \\
    \multicolumn{1}{c}{} & \multicolumn{1}{c}{} & Two-stream Fusion \cite{8} & 92.50\% & 65.40\% &   -    & 2016 \\
    \multicolumn{1}{c}{} & \multicolumn{1}{c}{} & SPN \cite{23}   & 94.60\% & 68.90\% &    -   & 2017 \\
    \multicolumn{1}{c}{} & \multicolumn{1}{c}{} & TCLSTA \cite{10}  & 94.00\% & 68.70\% &    -   & 2018 \\
    \specialrule{0em}{2.5pt}{2.5pt}
    \cline{2-7}
    \specialrule{0em}{2.5pt}{2.5pt}
    \multicolumn{1}{c}{} & \multicolumn{1}{c}{\multirow{9}[2]{*}{\rotatebox{90}{\textbf{RNN}}}} & LRCN \cite{13}  & 82.70\% &    -  &    -  & 2015 \\
    \multicolumn{1}{c}{} & \multicolumn{1}{c}{} & Beyond Short-Snippets \cite{12} & 88.20\% &   -   &     -  & 2015 \\
    \multicolumn{1}{c}{} & \multicolumn{1}{c}{} & Lattice-LSTM \cite{11}  & 93.60\% & 66.20\% &    -   & 2017 \\
    \multicolumn{1}{c}{} & \multicolumn{1}{c}{} &  Bi-LSTM \cite{14} & 91.21\% & 87.64\% &    -   & 2017 \\
    \multicolumn{1}{c}{} & \multicolumn{1}{c}{} & Db-LSTM \cite{15} & 97.30\% & 81.20\% &    -   & 2021 \\
    \multicolumn{1}{c}{} & \multicolumn{1}{c}{} & ShuttleNet \cite{21} & 95.40\% & 71.70\% &   -    & 2017 \\
    \multicolumn{1}{c}{} & \multicolumn{1}{c}{} & Attentional Pooling \cite{24} &    -   & 50.80\% &    -   & 2017 \\
    \multicolumn{1}{c}{} & \multicolumn{1}{c}{} & VideoLSTM \cite{26}& 79.60\% & 43.30\% &    -   & 2018 \\
    \multicolumn{1}{c}{} & \multicolumn{1}{c}{} & Spatio-temporal Attention \cite{25} & 87.11\% & 53.07\% &    -   & 2019 \\
   
    \specialrule{0em}{2.5pt}{2.5pt}
    \cline{2-7}
    \specialrule{0em}{2.5pt}{2.5pt}
    \multicolumn{1}{c}{} & \multicolumn{1}{c}{\multirow{9}[2]{*}{\rotatebox{90}{\textbf{3D CNN}}}} & C3D \cite{27}   & 82.30\% & 56.80\% & 59.50\% & 2015 \\
    \multicolumn{1}{c}{} & \multicolumn{1}{c}{} & I3D-Two Stream \cite{28} & 97.90\% & 80.20\% & 75.70\% & 2017 \\
    \multicolumn{1}{c}{} & \multicolumn{1}{c}{} & T3D \cite{34}   & 93.20\% & 63.50\% & 62.20\% & 2017 \\
    \multicolumn{1}{c}{} & \multicolumn{1}{c}{} &  R3D \cite{31}  & 94.50\% & 70.20\% & 65.10\% & 2018 \\
    \multicolumn{1}{c}{} & \multicolumn{1}{c}{} & (2+1)D \cite{33} & 97.30\% & 78.70\% & 75.40\% & 2018 \\
    \multicolumn{1}{c}{} & \multicolumn{1}{c}{} & SlowFast 8×8, R101 \cite{32} &    -   &     -  & 77.90\% & 2019 \\
    \multicolumn{1}{c}{} & \multicolumn{1}{c}{} & TSM \cite{30}   & 95.90\% & 73.50\% & 74.70\% & 2019 \\
    \multicolumn{1}{c}{} & \multicolumn{1}{c}{} & X3D-XL \cite{29} &    -   &    -   & 79.10\% & 2020 \\
    \specialrule{0em}{2.5pt}{2.5pt}
    \cline{2-7}
    \specialrule{0em}{2.5pt}{2.5pt}
    \multicolumn{1}{c}{} & \multicolumn{1}{c}{\multirow{6}[2]{*}{\rotatebox{90}{\textbf{Transformer}}}} & VTN \cite{neimark2021video}   &   -    &    -   & 79.80\% & 2021 \\
    \multicolumn{1}{c}{} & \multicolumn{1}{c}{} & ViViT \cite{41} &   -    &     -  & 84.80\% & 2021 \\
    \multicolumn{1}{c}{} & \multicolumn{1}{c}{} & Timesformer \cite{36} &    -   &    -   & 80.70\% & 2021 \\
    \multicolumn{1}{c}{} & \multicolumn{1}{c}{} & MTV-H (WTS) \cite{42} &    -   &    -   & 89.10\% & 2022 \\
    \multicolumn{1}{c}{} & \multicolumn{1}{c}{} & RegionViT \cite{39} &    -   &    -   & 77.60\% & 2022 \\
    \multicolumn{1}{c}{} & \multicolumn{1}{c}{} & RViT \cite{40}  &   -    &   -    & 81.50\% & 2022 \\
    \specialrule{0em}{2.5pt}{2.5pt}
    \cline{1-7}
    \specialrule{0em}{2.5pt}{2.5pt}
     \multicolumn{1}{c}{\multirow{4}[2]{*}{\shortstack{RGB\\and\\Audio}}} &  \multicolumn{1}{c}{\multirow{4}[2]{*}{}} & Wang et al. \cite{72} & 85.10\% &   -    &    -   & 2016 \\
    \multicolumn{1}{c}{} & \multicolumn{1}{c}{} & Long et al. \cite{75} & 94.60\% & 69.20\% & 79.40\% & 2018 \\
    \multicolumn{1}{c}{} & \multicolumn{1}{c}{} & AVSlowFast 8×8, R101 \cite{78} &   -    &   -    & 78.80\% & 2020 \\
    \multicolumn{1}{c}{} & \multicolumn{1}{c}{} & MM-VIT(Kinetics pretain) \cite{80} & 98.90\% &   -    &   -    & 2022 \\
    \specialrule{0em}{2.5pt}{2.5pt}
    \bottomrule
    \end{tabular}%
  \label{tab:addlabel}%
\end{table}%

As for 3D skeleton-based HAR, the performance based on 3D skeleton tada is shown in Table \ref{tab:skeleton} below, with experiments conducted by NTU-RGB+D and NTU-RGB+D 120 datasets. 

\begin{table}[h]
  \centering
  \caption{Performance of the skeleton-based HAR methods on NTU RGB+D and NTU RGB+D 120 datasets.}
    \label{tab:skeleton}
    \begin{tabular}{crcccccc}
    \toprule
    \multirow{2}{*}{\textbf{Modality}} &\multicolumn{2}{c}{\multirow{2}{*}{\textbf{Method}}} & \multicolumn{2}{c}{\textbf{NTU RGB+D}} & \multicolumn{2}{c}{\textbf{NTU RGB+D 120}} & \multirow{2}{*}{\textbf{Year}} \\ \cmidrule(lr){4-5}\cmidrule(lr){6-7}
          &       &       & Cross-Subject & Cross-View & Cross-Subject & Cross-Setup &  \\
    \midrule 
    \specialrule{0em}{2.5pt}{2.5pt}
    \multirow{19}{*}{\shortstack{3D\\Skeleton}} & \multicolumn{1}{c}{\multirow{6}{*}{\rotatebox{90}{\textbf{RNN}}}} & Two-stream RNN \cite{49} & 71.30\% & 79.50\% &-       &    -   & 2017 \\
          & \multicolumn{1}{c}{} & TS-LSTM \cite{50} & 74.60\% & 81.30\% &    -   &   -    & 2017 \\
          & \multicolumn{1}{c}{} & GCA-LSTM \cite{52} & 74.40\% & 82.80\% & 58.30\% & 59.20\% & 2017 \\
          & \multicolumn{1}{c}{} & IndRNN \cite{51} & 86.70\% & 93.70\% &    -   &     -  & 2018 \\
          & \multicolumn{1}{c}{} & MAN \cite{53} & 82.70\% & 93.20\% &  -     &    -   & 2018 \\
          & \multicolumn{1}{c}{} & VA-RNN \cite{79}  & 79.80\% & 88.90\% &    -   &    -   & 2019 \\
          \specialrule{0em}{2.5pt}{2.5pt}
           \cline{2-8}
           \specialrule{0em}{2.5pt}{2.5pt}
          & \multicolumn{1}{c}{\multirow{6}{*}{\rotatebox{90}{\textbf{CNN}}}} & JIM \cite{54}  & 73.40\% & 75.20\% &     -  &   -    & 2016 \\
          & \multicolumn{1}{c}{} & HCN \cite{55} & 86.50\% & 91.10\% &    -   &   -    & 2018 \\
          & \multicolumn{1}{c}{} & SkeleMotion \cite{57} & 76.50\% & 84.70\% & 67.70\% & 66.90\% & 2019 \\
          & \multicolumn{1}{c}{} & TSRJI \cite{58} & 73.30\% & 80.30\% & 67.90\% & 62.80\% & 2019 \\
          & \multicolumn{1}{c}{} &  GA \cite{56} & 82.90\% & 90.00\% &    -   &    -   & 2019 \\
          & \multicolumn{1}{c}{} & VA-CNN \cite{79}  & 88.70\% & 94.30\% &     -  & -      & 2019 \\
          \specialrule{0em}{2.5pt}{2.5pt}
           \cline{2-8}
           \specialrule{0em}{2.5pt}{2.5pt}
          & \multicolumn{1}{c}{\multirow{7}{*}{\rotatebox{90}{\textbf{GCN}}}} & ST-GCN \cite{59} & 81.50\% & 88.30\% &   -    &     -  & 2018 \\
          & \multicolumn{1}{c}{} & AS-GCN \cite{60} & 86.80\% & 94.20\% &    -   &   -    & 2019 \\
          & \multicolumn{1}{c}{} & CA-GCN \cite{61} & 83.50\% & 91.40\% &      - &   -    & 2020 \\
          & \multicolumn{1}{c}{} & Shift-GCN \cite{63} & 90.70\% & 96.50\% & 85.90\% & 87.60\% & 2020 \\
          & \multicolumn{1}{c}{} & SGN \cite{62}   & 89.00\% & 94.50\% & 79.20\% & 81.50\% & 2020 \\
          & \multicolumn{1}{c}{} & Sym-GNN \cite{65} & 90.10\% & 96.40\% &    -   &  -     & 2021 \\
          & \multicolumn{1}{c}{} & Info-GCN \cite{66} & 93.00\% & 97.10\% & 89.80\% & 91.20\% & 2022 \\
          \specialrule{0em}{2.5pt}{2.5pt}
    \bottomrule
    \end{tabular}%
  \label{tab:addlabel}%
\end{table}%

Additionally, with respect to multimodal action recognition using inertial sensor modality, Table \ref{tab:modality} also shows the performance on UTD-MHAD dataset.

\begin{table}[h]
  \centering
  \caption{Performance of the  multimodal HAR methods on UTD-MHAD dataset.\\SS:Subject-Specific,            SG:Subject-Generic}
    \label{tab:modality}
    \begin{tabular}{cccc}
    \toprule
   \textbf{ Modality} &\textbf{ Method} & \textbf{UTD-MHAD} &\textbf{ Year} \\
   \specialrule{0em}{2.5pt}{2.5pt}
    \midrule
    \specialrule{0em}{2.5pt}{2.5pt}
    \multirow{3}{*}{\shortstack{RGB \\and\\ Inertial Sensors}} & Wei et al. \cite{83} & 95.6\%(SG) & 2019 \\
          & \multirow{2}{*}{Wei et al. \cite{82}} & \multirow{2}{*}{\shortstack{91.3\%-smart TV gestures,\\
85.2\%-sports action(SG)}} & \multirow{2}{*}{2020} \\
          &       &       &  \\
          \specialrule{0em}{2.5pt}{2.5pt}
         \midrule   
         \specialrule{0em}{2.5pt}{2.5pt}
    \multirow{9}{*}{\shortstack{RGB-D \\and\\ Inertial Sensors}} &  Dawar et al. \cite{96} & 86.3\%(SS) & 2018 \\
          &  Dawar et al. \cite{99} & 89.20\%(SG) & 2018 \\
          &  Dawar et al. \cite{100} & 92.80\% & 2018 \\
          & Fuad et al. \cite{89} & 95.00\%(SG) & 2018 \\
          & Ahmad  et al. \cite{98} &  98.70\%(SS) & 2018 \\
          & Javed et al. \cite{97} & 98.30\%(SG) & 2019 \\
          & Imran et al. \cite{95} & 97.90\% & 2020 \\
          & MGAF \cite{102}  & 96.80\%(SS) & 2020 \\
          & ($\emph{M}^{2}$) fusion \cite{101} & 99.20\%(SS) & 2020 \\
          \specialrule{0em}{2.5pt}{2.5pt}
    \bottomrule
    \end{tabular}%
  \label{tab:addlabel}%
\end{table}%
 
As for EAR field, some of the more representative methods, which are employed to recognize action from the first view, are compared on the egocentric datasets in Table  \ref{tab:e}. The Acc in Table 5 are referred to as mean class accuracy, where the accuracy of each category is calculated separately, and the results of all classes were averaged.\\

\begin{table}[h]
  \centering
  \caption{Performance of the first-person action recognition  methods on egocentric datasets.\\Acc:mean class accuracy}
    \label{tab:e}
    \begin{tabular}{cccc}
    \toprule
   \textbf{ Dataset} & \textbf{Reference} & \textbf{Acc} &\textbf{ Year }\\
    \midrule
    \multirow{2}[0]{*}{Dogcentric} & Pooled Motion \cite{118}  & 73.00\% & 2015 \\
          & Kwon et al. \cite{121}   & 83.00\% & 2018 \\
            \midrule
    \multirow{5}[0]{*}{ADL} & Fathi et al. \cite{109}   & 49.80\% & 2011 \\
          & Pirsiavash et al. \cite{110}   & 36.70\% & 2012 \\
          & McCandless et al. \cite{111}   & 38.70\% & 2013 \\
          & Ego-ConvNet \cite{122}   & 37.58\% & 2016 \\
          & DCNN \cite{120}   & 55.20\% & 2016 \\
            \midrule
    \multirow{5}[0]{*}{GTEA gaze+} & Fathi et al. \cite{108}   & 47.70\% & 2011 \\
          & Li et al. \cite{116}   & 60.50\% & 2015 \\
          & Ma et al.  \cite{119}   & 66.40\% & 2016 \\
          &  Ego-ConvNet \cite{122}   & 68.50\% & 2016 \\
          & Ego-RNN \cite{125}   & 60.13\% & 2018 \\
            \midrule
    \multirow{10}[0]{*}{ EGTEA gaze+} & Li et al.\cite{116}   & 46.50\% & 2015 \\
          &  Ego-ConvNet \cite{122}   & 54.19\% & 2016 \\
          & Ego-RNN \cite{125}   & 52.40\% & 2018 \\
          & Li et al. \cite{131}   & 53.30\% & 2018 \\
          & Spatio-temporal Attention \cite{129}   & 60.54\% & 2019 \\
          & LSTA \cite{126}   & 53.00\% & 2019 \\
          & Lu et al. \cite{128}   & 46.84\% & 2019 \\
          & MCN \cite{132}   & 55.63\% & 2020 \\
          & Li et al.  \cite{133}   & 57.20\% & 2021 \\
          & APL \cite{136}   & 60.15\% & 2021 \\
    \bottomrule
    \end{tabular}%
  \label{tab:addlabel}%
\end{table}%

\section{Conclusion}
This paper firstly sorts out behavior recognition methods from the third-person view. For single modality, we introduce methods based on RGB video and 3D skeleton sequence respectively, which are the two most mainstream methods for single modal behavior recognition. In regard to multimodal input, the paper focuses on the multimodal fusion of visual sensors and non-visual sensors, including the multimodal fusion between RGB and audio modality, RGB and inertial sensors modality, RGB-D and inertial sensors modality. Given that the paper aims to provide a more comprehensive introduction to HAR for novices and researchers, thus we also investigate the action recognition methods from the first perspective in recent years. With respect to single modality, it mainly includes the traditional methods based on egocentric cue and the recognition methods based on deep learning. As for multimodal input, there are still relatively few studies, and paper briefly surveys the multimodal fusion between RGB and depth modality, RGB and audio modality, RGB and wearable inertial sensors modality. In addition, we also introduce the third-person view and egocentric datasets, and summarize the performance of the related methods on the corresponding datasets.

\bibliographystyle{plain}
\bibliography{bibliography.bib}

\end{document}